\documentclass[runningheads]{llncs}
\usepackage[T1]{fontenc}
\usepackage[latin9]{inputenc}
\usepackage{xcolor}
\usepackage{float}
\usepackage{varwidth}
\usepackage{amstext}
\usepackage{amssymb}
\usepackage{graphicx}
\PassOptionsToPackage{normalem}{ulem}
\usepackage{ulem}
\usepackage[pdfusetitle,
 bookmarks=false,
 breaklinks=false,pdfborder={0 0 1},backref=false,colorlinks=true]
 {hyperref}

\makeatletter

\providecommand{\tabularnewline}{\\}
\newenvironment{cellvarwidth}[1][t]
    {\begin{varwidth}[#1]{\linewidth}}
    {\@finalstrut\@arstrutbox\end{varwidth}}

\usepackage[misc]{ifsym}

\usepackage{amsfonts}
\usepackage{amssymb}

\usepackage{algorithm}
\usepackage{algorithmic}

\usepackage[table]{colortbl}
\definecolor{header_color}{rgb}{0.74,0.88,0.91}
\definecolor{even_color}{rgb}{0.9,0.9,0.9}
\definecolor{subheader_color}{rgb}{0.85,0.93,0.95}
\definecolor{childheader_color}{rgb}{1.0,0.93,0.87}

\author{
Tu Anh Hoang Nguyen\inst{1} \and
Dang Nguyen\inst{1} \and
Thuc Duy Le\inst{2} \and
Trung Le\inst{3} \and
Sunil Gupta\inst{1}
}

\authorrunning{T. Nguyen et al.}

\institute{
Applied Artificial Intelligence Initiative (A2I2), Deakin University, Australia\\
\email{\{joseph.nguyen,d.nguyen,sunil.gupta\}@deakin.edu.au}
\and
Adelaide University, Australia
\quad\quad
\textsuperscript{3}Monash University, Australia
}

\makeatother

\begin{document}
\title{LLM as Detector: An In-context Learning Approach for Tabular Anomaly
Detection}
\titlerunning{LLM as Detector: An In-context Learning Approach for Tabular Anomaly
Detection}
\maketitle
\begin{abstract}
Anomaly detection in tabular data is challenging because abnormal
samples often arise as violations of cross-feature dependencies rather
than simple marginal deviations. Existing detectors rely on geometric
or reconstruction signals, while prior LLM-based approaches mainly
fine-tune LLMs with normal samples or generate synthetic anomalies.
We propose \textbf{LLM-Detector}, a framework that utilizes the in-context
learning capacity of LLMs for structured, prompt-conditioned scoring
synthesis, enabling LLMs to derive anomaly detection logic from structured
normal-state knowledge. Specifically, normal training data are converted
into statistical summaries, causal dependencies, and distilled prototypes
that are organized into a prompt for code generation. The resulting
scoring engine evaluates statistical deviation, structural inconsistency,
and density-based abnormality then computes an anomaly score for each
test sample. We evaluate LLM-Detector on 24 tabular datasets, comparing
against 15 SOTA baselines. Results show consistent improvements across
both mixed-type and continuous-only settings. Moreover, this design
eliminates the need for LLM fine-tuning or neural network training,
reducing computational cost and enabling practical anomaly detection
in real-world tabular systems.

\keywords{Tabular Anomaly Detection \and Large Language Models \and In-context Learning}
\end{abstract}

\section{Introduction\label{sec:Introduction}}

\textit{Tabular Anomaly Detection} (TAD) is crucial in domains such
as finance, healthcare, and cybersecurity \cite{Huang2025}, where
detecting abnormal records can prevent costly failures or risks. In
practice, anomaly labels are rare and expensive, leading to highly
imbalanced settings where models are typically trained only on normal
data as \textit{unsupervised} task. Tabular data also presents intrinsic
challenges due to mixed feature types, heterogeneous scales, and the
absence of spatial or sequential inductive structure that many learning
methods rely on \cite{Mai2024}. Moreover, anomalies often appear
locally plausible but violate cross-feature dependencies or domain
mechanisms. As a result, abnormality in tabular data is frequently
structural and relational rather than purely numerical, making reliable
detection inherently challenging.

Most existing TAD methods are reconstruction-based approaches that
learn representations of normal data and detect anomalies through
reconstruction error \cite{schlegl2019f,Liu2021,Shenkar2022}. While
effective at capturing numeric deviation, these methods primarily
focus on reconstruction fidelity and rarely enforce explicit consistency
across features. More broadly, many detection paradigms treat anomalies
as distant points, relying heavily on density sparsity. However, anomalies
in tabular data often remain numerically plausible while violating
relational dependencies between variables. In addition, common pre-processing
for mixed-type data such as encoding categorical variables can distort
semantic meaning and weaken domain reasoning. Consequently, effective
TAD requires reasoning beyond purely numeric deviation \cite{Shi2022}.

\begin{figure}[th]
\centering
\includegraphics[width=0.8\textwidth]{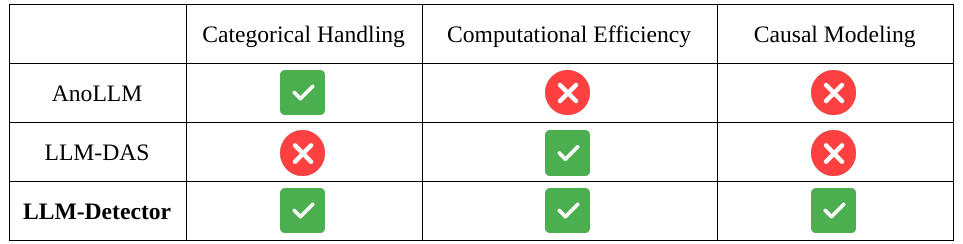}

\caption{\label{fig:Comparison-of-LLM-based}Comparison of LLM-based TAD methods
across categorical handling, computational efficiency, and causal
modeling.}
\end{figure}

Motivated by these limitations, recent studies have begun exploring
Large Language Models (LLMs) for TAD, with AnoLLM \cite{Tsai2025}
and LLM-DAS \cite{Ye2026} being the only representative LLM-based
approaches to date. AnoLLM directly applies LLMs as anomaly classifiers,
enabling mixed-type attributes to be represented through textual context
and semantic relations across features. However, this strategy suffers
from high computational cost and inefficient per-sample inference.
LLM-DAS mitigates this issue by generating hard anomalous samples
for conventional detectors, but its numerical backbones still encode
categorical attributes, risking semantic loss. As shown in Figure
\ref{fig:Comparison-of-LLM-based}, neither approach simultaneously
supports categorical handling, computational efficiency, and causal
modeling, motivating a deeper question: \textit{Can anomaly detection
instead emerge through in-context reasoning over structured normal-state
knowledge?}

To explore this question, we propose \textbf{LLM-Detector}, a framework
that leverages in-context learning without requiring LLM fine-tuning,
to synthesize an executable anomaly detection engine from structured
normal-state knowledge. LLM-Detector consists of two phases. First,
statistical profiles, causal relations, and representative prototypes
are extracted from normal training data and organized into a structured
prompt that induces the scoring logic. The LLM then generates a deterministic
program implementing this detection mechanism. Second, the synthesized
program evaluates unseen samples by aggregating statistical deviation,
structural inconsistency, and density-based abnormality signals to
produce anomaly scores. This design enables anomaly detection through
context-guided code generation rather than model training.

\textbf{Our contributions} are summarized as follows:

(1) We introduce LLM-Detector, a framework that treats LLMs as anomaly
detectors by inducing executable scoring logic through in-context
learning from structured normal-state knowledge, without LLM training/fine-tuning.

(2) We formalize a structured factorization of normal-state knowledge,
demonstrating that injecting statistical profiles, causal relations,
and representative prototypes into the prompt enables the generation
of causal-structure-guided scoring logic.

(3) We perform extensive evaluations on 24 diverse tabular benchmarks,
showing that the proposed framework achieves stronger detection performance
than SOTA statistical, deep learning, and recent LLM-based methods.

\section{Related Works\label{sec:Related-Works}}

\subsection{Unsupervised Anomaly Detection for Tabular Data}

Tabular anomaly detection (TAD) identifies atypical records in \textit{unlabeled}
datasets arising in many real-world domains. Since anomaly labels
are scarce and often unreliable, most practical approaches operate
in the \textit{unsupervised} setting. Existing methods typically detect
anomalies by modeling the distribution of normal data and identifying
samples that deviate from it. Classical approaches include proximity-based
methods such as KNN \cite{Ramaswamy2000}, Isolation Forest \cite{Liu2008},
PCA \cite{Shyu2003} and ECOD \cite{Li2022}. More recent work leverages
representation learning and deep models, including DeepSVDD \cite{Ruff2018}
and GOAD \cite{Bergman2020}, as well as reconstruction or generation-based
methods as REPEN \cite{Pang2018}, NeuTraL \cite{Qiu2021}, SLAD \cite{Xu2023},
and AnoGAN \cite{schlegl2019f}. More recently, CausalAno \cite{nguyen2026causal}
models causal dependencies to detect anomalies arising from mechanism
violations. Despite their differences, these approaches largely rely
on marginal deviation or density sparsity, which limits their ability
to capture relational inconsistencies among features.

\subsection{Large Language Models for Tabular Data Learning}

Recent studies have explored LLMs for learning from tabular data by
converting structured rows into textual representations and leveraging
contextual reasoning \cite{Borisov2023}. Beyond analysis tasks, LLMs
have also been applied to tabular data synthesis \cite{Borisov2023,Nguyen2024}.
More recently, AnoLLM \cite{Tsai2025} applies LLMs to detect anomalies
in tabular data through model fine-tuning on normal samples. Similarly, LLM-DAS
\cite{Ye2026} leverages LLMs to synthesize hard anomalous samples
to improve conventional detectors. However, these approaches mainly
use LLMs as predictors or data generators, leaving the role of LLMs
in constructing executable detection mechanisms largely unexplored.

\section{Framework\label{sec:Framework}}

\subsection{Problem Formulation}

Following previous works \cite{schlegl2019f,Livernoche2024,Tsai2025},
we investigate the problem of \textit{unsupervised} TAD, utilizing
a normal training set $D_{\text{normal}}=\{x_{i}\}_{i=1}^{N}$ composed
entirely of $N$ normal samples where $x_{i}\in\mathbb{R}^{d}$ consists
of numerical or categorical features. The task involves evaluating
a test set $D_{\text{test}}=\{(x'_{j},y'_{j})\}_{j=1}^{M}$, where
the label $y'_{j}\in\{0,1\}$ identifies a normal sample if $y'_{j}=0$
and an anomaly if $y'_{j}=1$. Our goal is to induce (without backpropagation
or model fine-tuning) an executable scoring function $s:\mathbb{R}^{d}\rightarrow\mathbb{R}$
from structured normal-state knowledge extracted from $D_{\text{normal}}$,
which assigns higher scores to anomalous instances in $D_{\text{test}}$.

\subsection{Proposed Method: LLM-Detector}

We propose LLM-Detector, a framework that harnesses the in-context
learning\textbf{ }capabilities of LLMs to automate the synthesis of
domain-specific anomaly detection engines. In our framework, in-context
learning is treated as a code synthesis mechanism. The LLM is not
fine-tuned; instead, it is conditioned on a compact prompt composed
of normal-state knowledge and scoring instructions. Specifically,
the prompt injects $p_{\text{stats}}$, $p_{\text{causal}}$, and
$p_{\text{distill}}$, together with explicit component-wise scoring
rules, into the LLM context. The LLM then transforms these four prompt
elements into an executable Python scoring engine that computes $S_{\text{outlier}}$,
$S_{\text{causal}}$, and $S_{\text{density}}$ for each test sample.
The model output is not a label prediction, but a deterministic algorithm
for calculating anomaly scores.

\begin{figure}[th]
\begin{centering}
\includegraphics[width=1\textwidth]{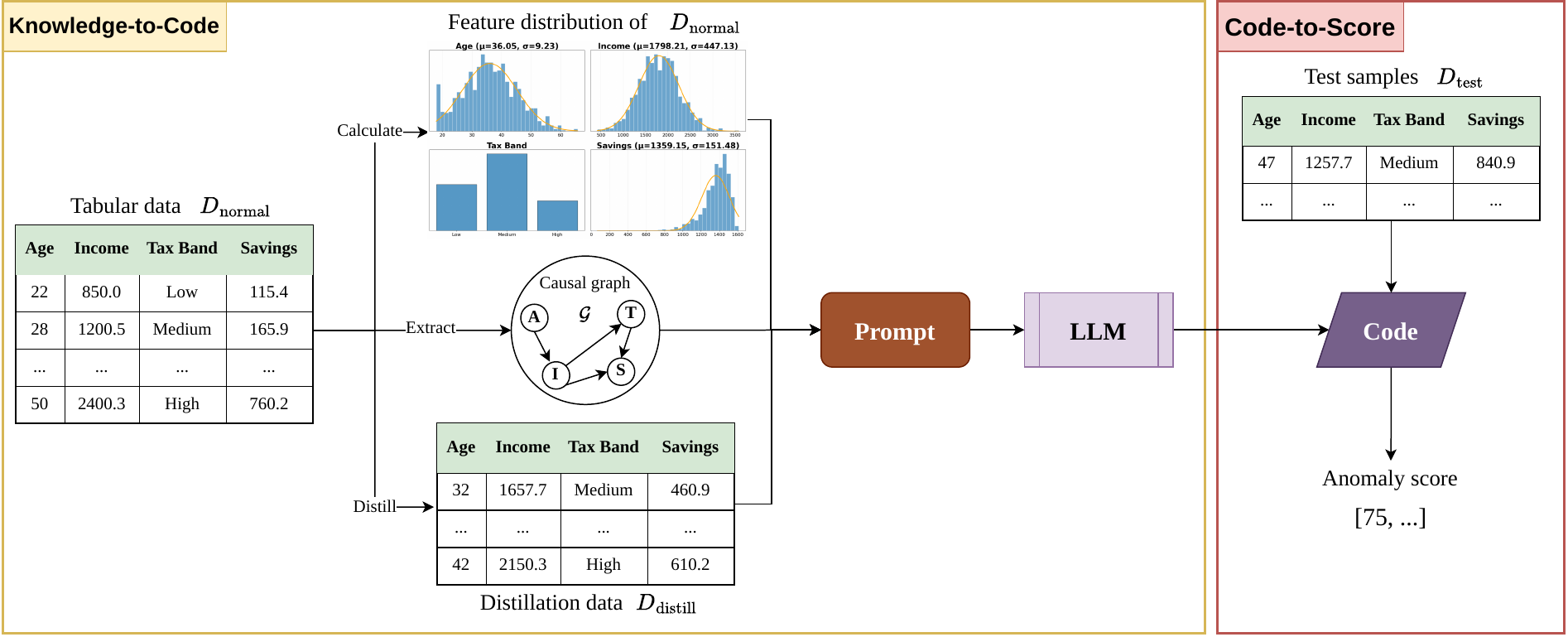}
\par\end{centering}
\caption{\label{fig:framework}Our framework \textbf{LLM-Detector}. It has
two phases: (1) \textbf{Knowledge-to-Code} extracts statistical profiles,
causal structure, and distilled prototypes from normal data to construct
a structured prompt, and (2) \textbf{Code-to-Score} generates a scoring
engine executed on test samples to compute anomaly scores.}
\end{figure}

As illustrated in Figure \ref{fig:framework}, our method operates
via a two-phase pipeline that cleanly decouples the construction of
a knowledge-dense prompt from the execution of a standalone scoring
engine. This design provides several practical advantages. Since the
LLM operates entirely through prompt-conditioned synthesis, LLM-Detector
does not require model training or fine-tuning. It also avoids heavy
pre-processing pipelines and allows the generated scoring engine to
be applied directly to $D_{\text{test}}$. Moreover, the prompt incorporates
structured knowledge derived from the normal state rather than directly
transmitting the full normal-sample dataset, thereby reducing data-leakage
risks. As a result, our method remains lightweight, privacy-aware,
and generalizable across domains.

\subsubsection{Knowledge-to-Code: Prompt Construction and Code Generation}

The primary objective of this phase is to convert the normal data
$D_{\text{normal}}$ into a compact \textquotedblleft normal-state
knowledge package\textquotedblright . By utilizing the LLM in a \textbf{zero-shot
capacity}, we drive the synthesis of an executable scoring engine
without altering any model weights or requiring specialized GPU clusters
for training. To formalize this zero-shot synthesis process, we structure
the code-generation request into three functional components:
\begin{equation}
p_{\text{code}}=p_{\text{description}}+p_{\text{objective}}+p_{\text{requirements}}\label{eq:prompt}
\end{equation}

\paragraph{\textbf{Description $p_{\text{description}}$}}

The description component encodes dataset-derived normal-state knowledge
that the generated scoring engine must operationalize. It encodes
dataset-derived normal-state knowledge extracted from $D_{\text{normal}}$,
defining the informational foundation upon which the scoring logic
is constructed.

The description component is further decomposed as:
\begin{equation}
p_{\text{description}}=p_{\text{stats}}+p_{\text{causal}}+p_{\text{distill}}\label{eq:description}
\end{equation}

This factorization makes explicit how each score term is grounded
in a corresponding knowledge source extracted from $D_{\text{normal}}$.

\textbf{(i) Statistical profile  $p_{\text{stats}}$}

We summarize the marginal behavior of each feature into a statistical
profile $p_{\text{stats}}$, fixing the expected feature identities
and dimensionality for the scoring engine. For each feature $X_{i}$,
we define a per-feature profile $\phi_{i}$ and store either numerical
statistics with empirical bounds or categorical empirical probabilities,
as specified in Equation \ref{eq:stat_profile}.
\begin{equation}
\phi_{i}=\left\{ \begin{array}{ll}
(\mu_{i},\sigma_{i}^{2},[min_{i},max_{i}]) & \text{if }X_{i}\text{ is numerical}\\[4pt]
\{\pi_{c}\}_{c\in\mathcal{C}_{i}} & \text{if }X_{i}\text{ is categorical}
\end{array}\right.,\label{eq:stat_profile}
\end{equation}
where $\mu_{i}$, $\sigma_{i}^{2}$, $min_{i}$, and $max_{i}$ are
mean, variance, minimum value, and maximum value of the continuous
feature $X_{i}$ while $\pi_{c}$ is a proportion of a category $c$
in the domain $C_{i}$ of a categorical feature $X_{i}$.

\textbf{(ii) Causal knowledge $p_{\text{causal}}$}

We represent the feature dependency in the normal data $D_{\text{normal}}$
by a causal graph (DAG) $\mathcal{G}$ over features $X=\{X_{1},\ldots,X_{d}\}$,
where each node $X_{i}$ is associated with a parent set $\mathrm{PA}(X_{i})$
capturing its conditional relations. We apply the Peter-Clark (PC)
algorithm \cite{Spirtes2000} to an encoded and normalized representation
of $D_{\text{normal}}$ to estimate $\mathcal{G}$, using an adaptive
independence-testing strategy to avoid degenerate outputs. These parent-child
relations expose cross-feature constraints, helping the LLM induce
implicit rules for code-based scoring.

\textbf{(iii) Distilled samples $p_{\text{distill}}$}

\begin{figure}[t]
\begin{centering}
\includegraphics[width=1\textwidth]{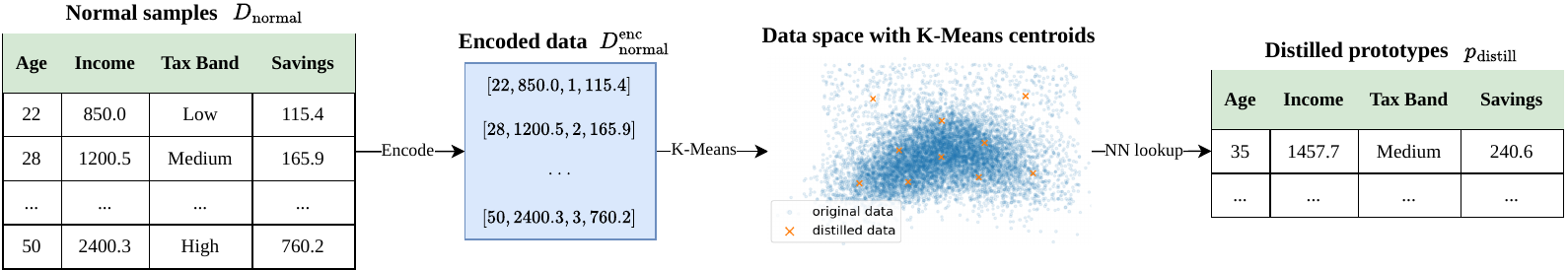}
\par\end{centering}
\caption{\label{fig:Prototype-distillation}Prototype distillation via K-Means
and nearest-neighbor (NN) mapping.}
\end{figure}

Since in-context learning is constrained by the finite context window
of LLMs, which limits the number of demonstrations that can be included
in a prompt \cite{brown2020language}, we distill $D_{\text{normal}}$
into $N_{\text{distill}}$ representative prototypes to provide a
compact yet informative summary of the normal data distribution. As
illustrated in Figure \ref{fig:Prototype-distillation}, we first
encode and normalize the normal set into a consistent continuous space
$D_{\text{normal}}^{\text{enc}}$, perform K-Means clustering to obtain
$N_{\text{distill}}$ centroids, then map each centroid to its nearest
normal instance via the nearest-neighbor (NN) operator. We perform
the mapping step to make sure each prototype is a valid real normal
sample. The resulting subset $p_{\text{distill}}$ serves as geometry
anchors for covariance estimation and low-density assessment, enabling
Mahalanobis-style multivariate deviation checks in the scoring engine.
Formally, we derive $p_{\text{distill}}$ via:
\begin{equation}
p_{\text{distill}}=\mathrm{NN}\!\big(\mathrm{K\text{-}Means}(D_{\text{normal}}^{\text{enc}},N_{\text{distill}}),\,D_{\text{normal}}^{\text{enc}}\big)\label{eq:p_distill}
\end{equation}

\paragraph{\textbf{Objective $p_{\text{objective}}$}}

Building upon the injected structured normal-state knowledge, the
LLM is instructed to generate an end-to-end Python scoring engine
rather than labels or narrative analysis. The output must translate
the provided knowledge into explicit computations. The engine maps
each \textit{unseen} test sample $x$ to a continuous anomaly score
in $[0,100]$, where $0$ denotes perfectly normal behavior and $100$
denotes extreme abnormality. The output score is calibrated and comparable
across samples to keep the logic deterministic and grounded in the
injected normal-state knowledge.

\paragraph{\textbf{Requirements $p_{\text{requirements}}$}}

To ensure faithful use of the injected normal-state knowledge, we
constrain both score computation and program structure:
\begin{itemize}
\item \textbf{Score decomposition.} The engine shall compute a single continuous
anomaly score by aggregating scores of three prescribed terms, as
specified in $p_{\text{description}}$.
\item \textbf{Budgeted components.} Each term shall be allocated a fixed
score budget to control its contribution and prevent dominance by
any single component.
\item \textbf{Bounded output.} The aggregated score shall be properly bounded
within $[0,100]$ to preserve calibration and comparability across
samples.
\item \textbf{Callable interface.} The code shall define $\texttt{evaluate\_anomalies(new\_samples)}$
that takes a pandas DataFrame and returns an $\texttt{anomaly\_score}$.
\item \textbf{Embedded knowledge.} All injected normal-state knowledge shall
be embedded directly to ensure deterministic test-time execution.
\end{itemize}
After assembling $p_{\text{code}}$, the complete prompt (containing
description, objective, and requirements) is submitted to the LLM
as a code-generation request. The LLM returns executable Python code
that directly implements the scoring logic based on the provided statistics,
causal graph, and representative samples. The generated program contains
explicit mathematical operations rather than placeholders or learned
parameters. As no fine-tuning is performed, the LLM serves purely
as a deterministic code synthesizer. The full prompt template is provided
in \textbf{Appendix 4}.

\subsubsection{Code-to-Score: Test-Time Inference and Anomaly Scoring}

At test time, the synthesized Python scoring engine is executed on
an unlabeled feature matrix $X_{\text{test}}$ that follows the same
feature schema and types defined in the injected normal-state knowledge.
Note that no information from $D_{\text{test}}$ is used to modify
the scoring logic.

For each test sample $x\in D_{\text{test}}$, the final anomaly score
is computed as the aggregation of three score-components:
\begin{equation}
S(x)=S_{\text{outlier}}(x)+S_{\text{causal}}(x)+S_{\text{density}}(x).\label{eq:scoring}
\end{equation}

\textbf{Outlier score-component $S_{\text{outlier}}$:} Guided by
the statistical profile in Equation \ref{eq:stat_profile}, this component
measures marginal outlierness. For a continuous feature $X_{i}$,
the engine first computes a standardized deviation score $z_{i}=\frac{|x_{i}-\mu_{i}|}{\sigma_{i}}$,
where $\mu_{i}$ and $\sigma_{i}$ are obtained from the normal-state
profile. Larger deviations receive progressively higher penalties,
while values outside the empirical range $[min_{i},max_{i}]$ are
strongly penalized as out-of-support observations. For a categorical
feature, the penalty is determined by its empirical normal probability
$\pi_{c}$, with rare or unseen categories receiving high risk scores.
The resulting per-feature penalties are aggregated and rescaled to
a fixed $S_{\text{outlier}}$ budget, yielding a bounded marginal-outlierness
score.

\textbf{Causal violation score-component $S_{\text{causal}}$:} This
component checks whether a test sample breaks the dependency structure
encoded by the discovered DAG. For each child feature $X_{i}$, the
engine first derives an abnormality indicator $a_{i}(x_{i})$ from
its feature-level deviation, where larger values indicate stronger
evidence that $X_{i}$ is abnormal. The parent condition is summarized
as $a_{\mathrm{PA}(i)}(x)=\max_{X_{j}\in\mathrm{PA}(X_{i})}a_{j}(x_{j})$.
A causal violation is detected when $a_{i}(x_{i})>\tau_{c}$ and $a_{\mathrm{PA}(i)}(x)\leq\tau_{p}$,
where $\tau_{c}$ and $\tau_{p}$ denote the child-abnormality and
parent-normality thresholds, respectively. This means that the child
enters an abnormal state while its causal parents remain normal, indicating
a broken conditional dependency. Such violations receive structural
penalties, which are aggregated and rescaled to the fixed S$_{\text{causal}}$
budget.

\textbf{Density score-component $S_{\text{density}}$}: This component
checks whether a test sample lies in a low-density region of the normal-state
distribution. Using the representative normal samples, the engine
estimates the empirical mean vector $\mu$ and covariance matrix $\Sigma$
over continuous features, then computes the Mahalanobis distance $d_{M}(x)=\sqrt{(x-\mu)^{\top}\Sigma^{-1}(x-\mu)}$.
This distance captures correlation-aware multivariate deviation from
the normal manifold. Larger distances receive higher penalties, which
are scaled and bounded within the fixed $S_{\text{density}}$ budget.

Each score-component $S_{\text{outlier}}$, $S_{\text{causal}}$,
and $S_{\text{density}}$ operates within a predefined score allocation,
ensuring that statistical deviation, causal inconsistency, and density-based
abnormality contribute in a controlled and interpretable manner. The
aggregated score $S(x)$ is subsequently bounded to a fixed range
to guarantee comparability across samples, datasets, and experimental
runs.

The engine outputs a continuous anomaly score $S(x)$ between 0 and
100 for each test sample, where larger scores indicate stronger abnormality.
The labels (0 for normal and 1 for anomalous) are used only for evaluation
(e.g., computing AUC-ROC) and never used for score construction, hyper-parameter
tuning, or rule design, preserving the unsupervised, in-context learning
setting.

A critical advantage of LLM-Detector is that the LLM is not merely
instantiating a static template. Unlike manually engineered detectors
with fixed, universally applied scoring logic, the LLM dynamically
synthesizes dataset-specific anomaly programs conditioned on the provided
statistical, structural, and prototype knowledge. By processing mixed-type
attributes and causal graphs through textual context, the LLM adapts
categorical handling, feature interactions, and component calibration
to the semantic characteristics of each unique dataset--bridging
the gap between raw statistical data and executable, causal-aware
detection rules without requiring handcrafted algorithm design.

\section{Experiments\label{sec:Experiments}}

\subsection{Experimental Setups }

\subsubsection{Datasets}

We evaluate our method on 24 anomaly detection datasets, following
standard protocols in prior works \cite{Livernoche2024,Tsai2025,Ye2026}.
The benchmark combines datasets from Outlier Detection DataSets (ODDS)
\cite{Rayana2016}, Anomaly Detection Benchmark (ADBench) \cite{Han2022},
and Kaggle.

Overall, our benchmark datasets include 12 mixed-type datasets (with
categorical and continuous features) and 12 continuous-only datasets.
These datasets span diverse domains--such as healthcare, finance,
cybersecurity, and social sciences--and vary in both dimensionality
and scale, covering settings from small, low-dimensional data to larger,
high-dimensional scenarios. Table \ref{tab:mix-data} shows details
of 12 mixed-type datasets while the characteristics of 12 continuous-only
datasets are reported in \textbf{Appendix 1}.

\begin{table}[th]
\centering
\caption{\label{tab:mix-data}Statistics for 12 mixed-type datasets. $d_{cat}$,
$d_{con}$, and $N_{a}$ denote the numbers of categorical features,
continuous features, and anomalies. \textit{Lymp} (Lymphography),
\textit{ACD} (Cybersecurity), and \textit{Fraud} (Fraud ecommerce)
are from \cite{Rayana2016,francesco_capurso_2024,grover2022fraud}
while \textit{SPD} (Spyware-Attacks), \textit{DAMRE} (Damage-Report),
\textit{OS} (OS-Kernel), \textit{SMD} (Smart-Meter), and \textit{Vifd}
(Vehicle insurance) are from Kaggle.}

\begin{tabular}{|l|r|r|r|r|r|}
\hline 
\rowcolor{header_color}Dataset & $N$ & $d$ & $d_{cat}$ & $d_{con}$ & $N_{a}$\tabularnewline
\hline 
\hline 
\rowcolor{even_color}\textit{ACD} & 10,000 & 4 & 3 & 1 & 490 (5\%)\tabularnewline
\hline 
\textit{Bank} & 41,188 & 10 & 10 & 0 & 4,640 (11\%)\tabularnewline
\hline 
\rowcolor{even_color}\textit{CMC} & 1,473 & 8 & 8 & 0 & 29 (2\%)\tabularnewline
\hline 
\textit{DAMRE} & 1,000 & 5 & 2 & 3 & 100 (10\%)\tabularnewline
\hline 
\rowcolor{even_color}\textit{Fraud} & 151,112 & 8 & 6 & 2 & 14,151 (9\%)\tabularnewline
\hline 
\textit{Lymp} & 148 & 18 & 15 & 3 & 6 (4\%)\tabularnewline
\hline 
\rowcolor{even_color}\textit{NHIS} & 4,388 & 4 & 2 & 2 & 56 (1\%)\tabularnewline
\hline 
\textit{OS} & 1,000 & 5 & 2 & 3 & 91 (9\%)\tabularnewline
\hline 
\rowcolor{even_color}\textit{Seismic} & 2,584 & 18 & 4 & 14 & 170 (7\%)\tabularnewline
\hline 
\textit{SMD} & 5,000 & 7 & 2 & 5 & 250 (5\%)\tabularnewline
\hline 
\rowcolor{even_color}\textit{SPD} & 1,000 & 12 & 9 & 3 & 764 (76\%)\tabularnewline
\hline 
\textit{Vifd} & 15,420 & 32 & 24 & 8 & 923 (6\%)\tabularnewline
\hline 
\end{tabular}
\end{table}

\subsubsection{Evaluation Metrics}

To maintain consistency with other TAD papers, we adopt the same data
partitioning strategy commonly used in the literature \cite{Bergman2020,Livernoche2024,Tsai2025}.
Specifically, 50\% of the normal samples are randomly selected to
form the normal set $D_{\text{normal}}$, while the remaining 50\%
of normal samples are merged with all available anomalies to construct
the test set $D_{\text{test}}$.

For evaluation, we use the AUC-ROC metric, following the standard
reporting practices \cite{Han2022,Livernoche2024,Tsai2025}. This
metric assesses the model\textquoteright s ability to distinguish
between normal and anomalous samples by measuring ranking performance
across all possible decision thresholds. We also report the F1-score
in \textbf{Appendix 3}.

\subsubsection{Baseline Methods}

We compare our approach LLM-Detector against 15 SOTA methods spanning
classical algorithms, deep representation learning, self-supervised
techniques, generative models, and recent LLM-based detectors. The
compared methods include IForest \cite{Liu2008}, KNN \cite{Ramaswamy2000},
PCA \cite{Shyu2003}, ECOD \cite{Li2022}, DeepSVDD \cite{Ruff2018},
REPEN \cite{Pang2018}, RCA \cite{Liu2021}, SLAD \cite{Xu2023},
GOAD \cite{Bergman2020}, NeuTraL \cite{Qiu2021}, ICL \cite{Shenkar2022},
DTE \cite{Livernoche2024}, AnoGAN \cite{schlegl2019f}, AnoLLM \cite{Tsai2025},
and LLM-DAS \cite{Ye2026}.

For fairness and reproducibility, we rely on their widely used public
implementations. Thirteen baselines (from IForest to AnoGAN) are executed
through the PyOD library, while AnoLLM and LLM-DAS are evaluated using
their released source codes. All methods are trained and tested under
identical data splits, pre-processing steps, and evaluation metrics.

\subsubsection{Implement Details}

For the code generation phase, we utilize Gemini-3.0 to synthesize
the executable Python scoring engine. To ensure a representative yet
compact knowledge package, the number of distilled samples is fixed
at $N_{\text{distill}}=\text{min}(100,N)$. This distillation strategy
captures the multivariate geometry while maintaining prompt efficiency.
To mitigate the inherent stochasticity of the LLM and ensure performance
stability, each experiment is repeated three times using different
random seeds. We report the average performance result along with
its standard deviation.

\subsection{Results and Analysis}

\subsubsection{Results on Mixed-type Datasets}

Table \ref{tab:mix-result} reports AUC-ROC on 12 mixed-type datasets.
Overall, LLM-based methods consistently outperform classical statistical
and deep representation-based approaches, achieving consistently higher
average AUC-ROC scores. This trend highlights the advantage of textual
representation for modeling heterogeneous mixed-type tabular data,
where capturing interactions between continuous and categorical variables
is critical.

LLM-based methods such as AnoLLM (0.6972) and LLM-DAS (0.6900) demonstrate
clear improvements over non-LLM baselines. However, our method LLM-Detector
further advances this line of work, achieving the highest average
AUC-ROC of 0.7407 (5\% better than the second-best method AnoLLM).
This margin confirms that explicitly structuring statistical, causal,
and representative normal-state knowledge within the prompt leads
to more robust and stable performance across mixed-type datasets.
Notably, our LLM-Detector does not fine-tune/train LLMs, running much
faster than AnoLLM as shown in a later ablation study.

\begin{table}[th]
\caption{\label{tab:mix-result}AUC-ROC scores for all methods on 12 mixed-type
datasets. \textbf{Bold} and \uline{underline} indicate the best and
second-best methods. Standard deviations are reported in \textbf{Appendix
2}.}

\resizebox{\textwidth}{!}{
\centering{}%
\begin{tabular}{|l|cccccccccccc|c|}
\hline 
\rowcolor{header_color} & ACD & Bank & CMC & DAMRE & Fraud & Lymp & NHIS & OS & Seismic & SMD & SPD & Vifd & \textcolor{red}{AVG}\tabularnewline
\hline 
\hline 
Iforest & 0.5030 & 0.5118 & 0.5061 & 0.6901 & 0.4800 & 0.7050 & 0.6573 & 0.7716 & 0.7058 & 0.8127 & 0.4837 & 0.5067 & \textcolor{red}{0.6112}\tabularnewline
\rowcolor{even_color}KNN & 0.4943 & 0.5124 & \uline{0.5779} & 0.9250 & 0.5005 & 0.8705 & 0.6748 & 0.9330 & \textbf{0.7445} & 0.9237 & 0.4919 & 0.5227 & \textcolor{red}{0.6809}\tabularnewline
PCA & 0.5059 & 0.5152 & 0.5101 & 0.8857 & 0.4538 & 0.8560 & 0.6498 & 0.8475 & 0.7041 & 0.9340 & 0.4786 & 0.4931 & \textcolor{red}{0.6528}\tabularnewline
\rowcolor{even_color}ECOD & 0.5086 & 0.5134 & 0.5394 & 0.8442 & 0.4631 & 0.8545 & 0.6537 & 0.7686 & 0.6980 & 0.9114 & 0.4915 & 0.5028 & \textcolor{red}{0.6458}\tabularnewline
DeepSVDD & 0.4774 & 0.4996 & 0.5715 & 0.9162 & 0.5015 & 0.8685 & 0.6626 & 0.7957 & 0.6939 & 0.6864 & 0.4848 & 0.4993 & \textcolor{red}{0.6381}\tabularnewline
\rowcolor{even_color}GOAD & 0.4969 & 0.5139 & 0.4833 & 0.8726 & 0.5001 & 0.8764 & 0.7151 & 0.9117 & 0.7193 & 0.1545 & 0.4910 & 0.4991 & \textcolor{red}{0.6028}\tabularnewline
ICL & \textbf{0.5294} & 0.5185 & 0.5468 & \uline{0.9293} & 0.4760 & 0.9077 & \uline{0.8501} & \textbf{0.9613} & 0.7142 & 0.8570 & 0.4805 & \uline{0.5579} & \textcolor{red}{0.6941}\tabularnewline
\rowcolor{even_color}RCA & 0.5008 & 0.5138 & 0.5029 & 0.9146 & 0.5014 & 0.9210 & 0.6613 & 0.8650 & \uline{0.7323} & 0.9260 & 0.4940 & 0.5294 & \textcolor{red}{0.6719}\tabularnewline
SLAD & 0.5010 & 0.5315 & 0.4924 & 0.8039 & 0.5242 & 0.9397 & 0.8252 & 0.8677 & 0.7101 & 0.4732 & 0.4495 & \textbf{0.5645} & \textcolor{red}{0.6402}\tabularnewline
\rowcolor{even_color}NeuTral & 0.4925 & 0.5113 & 0.5457 & 0.8271 & 0.5037 & 0.7700 & 0.6733 & 0.7920 & 0.6740 & 0.5075 & \textbf{0.5361} & 0.4997 & \textcolor{red}{0.6111}\tabularnewline
DTE & 0.5106 & 0.5901 & 0.5443 & 0.8940 & 0.5000 & 0.8654 & 0.4900 & \uline{0.9409} & 0.7047 & 0.8096 & 0.4814 & 0.5422 & \textcolor{red}{0.6561}\tabularnewline
\rowcolor{even_color}REPEN & 0.5037 & 0.5127 & \textbf{0.5971} & 0.7840 & 0.5022 & 0.8552 & 0.6699 & 0.9317 & 0.7295 & 0.7877 & 0.4788 & 0.5096 & \textcolor{red}{0.6552}\tabularnewline
AnoGAN & 0.5039 & 0.5286 & 0.4298 & 0.6450 & 0.4551 & 0.9898 & 0.4747 & 0.6945 & 0.6503 & 0.7336 & 0.5027 & 0.5260 & \textcolor{red}{0.5945}\tabularnewline
\rowcolor{even_color}AnoLLM & 0.5074 & \textbf{0.6479} & 0.5301 & 0.9278 & \uline{0.6043} & \uline{0.9945} & 0.5342 & 0.9210 & 0.7382 & 0.9166 & 0.4880 & 0.5564 & \textcolor{red}{\uline{0.6972}}\tabularnewline
LLM-DAS & 0.5015 & 0.6172 & 0.5083 & 0.8806 & 0.4538 & 0.9914 & 0.8020 & 0.8597 & 0.7006 & \uline{0.9268} & 0.4999 & 0.5381 & \textcolor{red}{0.6900}\tabularnewline
\rowcolor{even_color}LLM-Detector & \uline{0.5189} & \uline{0.6448} & 0.5475 & \textbf{0.9300} & \textbf{0.7100} & \textbf{0.9949} & \textbf{0.8848} & 0.9105 & 0.7103 & \textbf{0.9714} & \uline{0.5312} & 0.5339 & \textbf{\textcolor{red}{0.7407}}\tabularnewline
\hline 
\end{tabular}}
\end{table}

\subsubsection{Results on Continuous Datasets}

\begin{figure}
\begin{centering}
\includegraphics[width=0.8\textwidth]{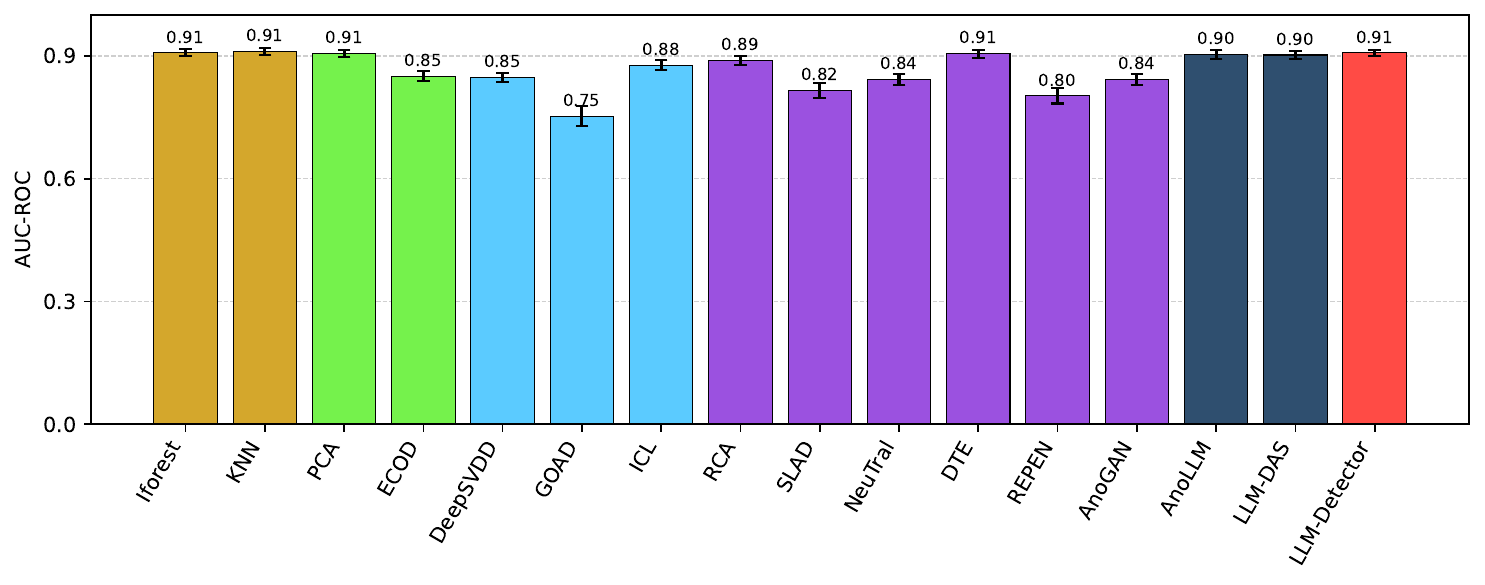}
\par\end{centering}
\caption{\label{fig:continuous-results}AUC-ROC comparison on 12 continuous
datasets across all 15 methods. The color scheme is: \textcolor{yellow}{yellow}
(proximity-based), \textcolor{green}{green} (distribution-based),
\textcolor{blue}{blue} (boundary-based), \textcolor{violet}{purple}
(reconstruction/generation-based), \textcolor{teal}{navy} (LLM-based),
\textcolor{red}{red} (ours).}
\end{figure}

Figure \ref{fig:continuous-results} shows that on continuous datasets,
the overall performance gap is narrower than in mixed-type settings,
with several classical and deep methods achieving strong AUC-ROC scores
around 0.85-0.91. Nevertheless, LLM-based methods remain competitive,
and our LLM-Detector attains the top-tier performance (0.91). Compared
with AnoLLM and LLM-DAS, LLM-Detector is slightly better, demonstrating
that the structured Knowledge-to-Code design generalizes effectively
to fully continuous feature spaces.

Across both mixed-type and continuous datasets, our LLM-Detector achieves
the highest average AUC-ROC with a deterministic scoring engine designed
from structured statistical, causal, and distilled knowledge, enabling
robust anomaly detection through principled in-context learning code
generation (without detector parameter training or LLM fine-tuning).

\subsection{Ablation Study}

\subsubsection{Knowledge in the Description Component $p_{\text{description}}$}

\begin{figure}
\begin{centering}
\includegraphics[width=0.8\textwidth]{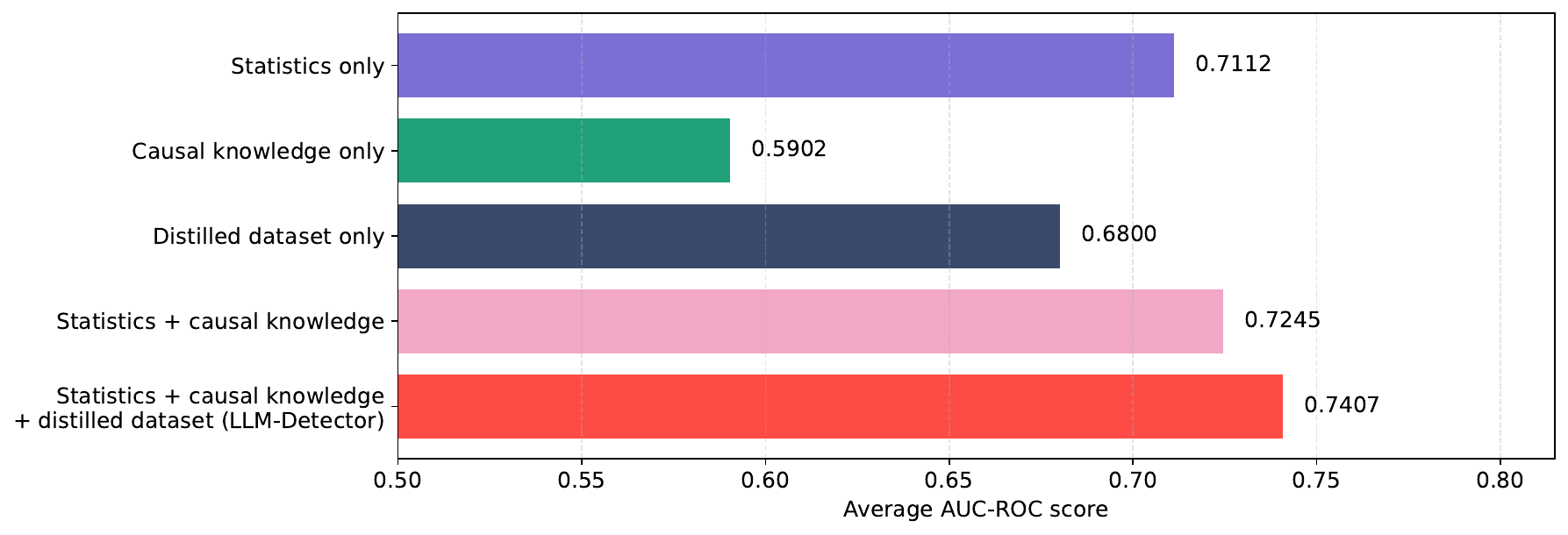}
\par\end{centering}
\caption{\label{fig:components}Ablation study of three types of knowledge
in the description component on 12 mixed-type datasets.}
\end{figure}

We study the impact of three types of knowledge in $p_{\text{description}}$
over the 12 mixed-type datasets (Figure \ref{fig:components}). Using
causal knowledge only yields the lowest performance (0.5902), which
is expected since the LLM receives only structural relations without
sufficient information about feature distributions or even feature
types (continuous vs. categorical). Without marginal context, the
model cannot reliably quantify abnormality. In contrast, statistics
achieves the best single-component result (0.7112), as marginal deviation
provides a direct and stable anomaly signal.

Combining statistics information with causal knowledge improves AUC-ROC
to 0.7245, suggesting that anomalies are not only distant from marginal
norms but can also violate the dependency mechanisms. Adding distilled
samples further captures sample-level multivariate structure beyond
feature-wise summaries, enabling covariance estimation and Mahalanobis-distance
geometry checks. Together, these three knowledge achieve the highest
average AUC-ROC of 0.7407.

\subsubsection{LLM Backbones}

\begin{figure}[th]
\begin{centering}
\includegraphics[width=0.8\textwidth]{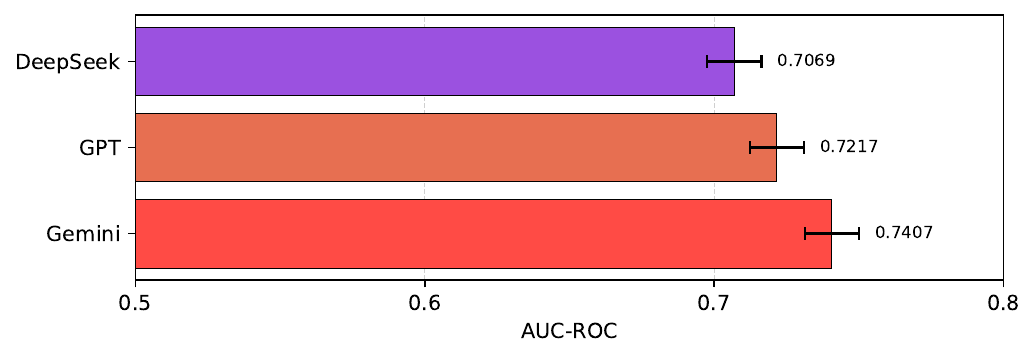}
\par\end{centering}
\caption{\label{fig:llm_backbones}Sensitivity to three LLM backbones (DeepSeek,
GPT, and Gemini) over 12 mixed-type datasets.}
\end{figure}

To examine the impact of different LLM foundation models, we evaluate
LLM-Detector using DeepSeek V3.2, GPT-5.2, and our default backbone
Gemini-3.0. Figure \ref{fig:llm_backbones} reports the average AUC-ROC
across 12 mixed-type datasets. All backbones yield competitive results,
confirming that the proposed Knowledge-to-Code framework is model-agnostic.
Namely, DeepSeek V3.2 achieves 0.7069, GPT-5.2 improves to 0.7217,
and Gemini-3.0 attains the highest performance at 0.7407. We attribute
the Gemini-3.0\textquoteright s superior results to its stronger long-context
reasoning and more reliable code synthesis \cite{rahman2025comparative},
which better preserve the structured statistical, causal, and distilled
knowledge embedded in the prompt. Overall, stronger backbone reasoning
improves detection accuracy, while the stable performance across models
shows that the main gains come from the Knowledge-to-Code design rather
than backbone-specific selection.

\subsubsection{The Number of Distilled Samples}

\begin{figure}[th]
\centering{}\includegraphics[width=0.8\textwidth]{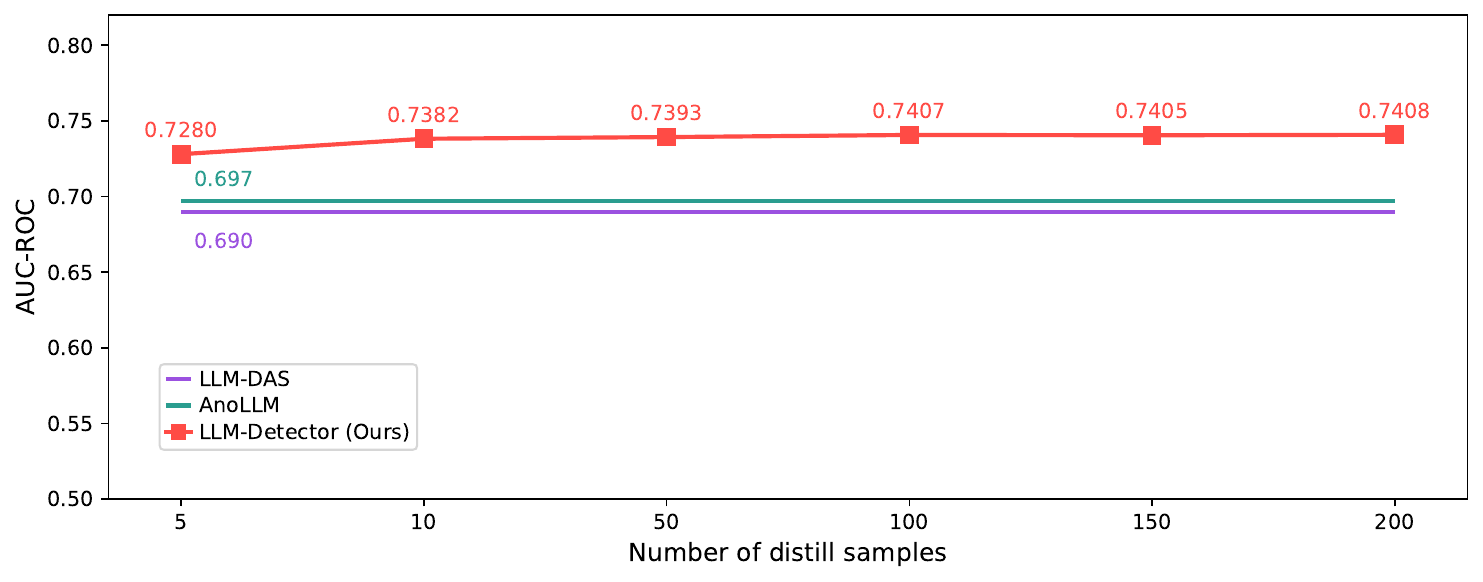}\caption{\label{fig:num_distill}Sensitivity to the number of distilled samples
over 12 mixed-type datasets.}
\end{figure}

We analyze the effect of varying the number of distilled normal prototypes
$N_{\text{distill}}$ on the 12 mixed-type datasets. Figure \ref{fig:num_distill}
reports the average AUC-ROC as $N_{\text{distill}}$ increases from
5 to 200. With fewer than 100 distilled samples, performance remains
noticeably lower, indicating that insufficient prototypes fail to
adequately capture the multivariate geometry of the normal state.
Performance improves steadily as $N_{\text{distill}}$ increases to
100, after which the gains stabilize. Using 150 or 200 distilled samples
yields nearly identical AUC-ROC to 100, suggesting diminishing returns.
More importantly, our method LLM-Detector consistently outperforms
AnoLLM and LLM-DAS across $N_{\text{distill}}$ values.

\subsubsection{Effect of Causal Discovery Methods}

\begin{figure}[H]
\centering
\includegraphics[width=0.8\textwidth]{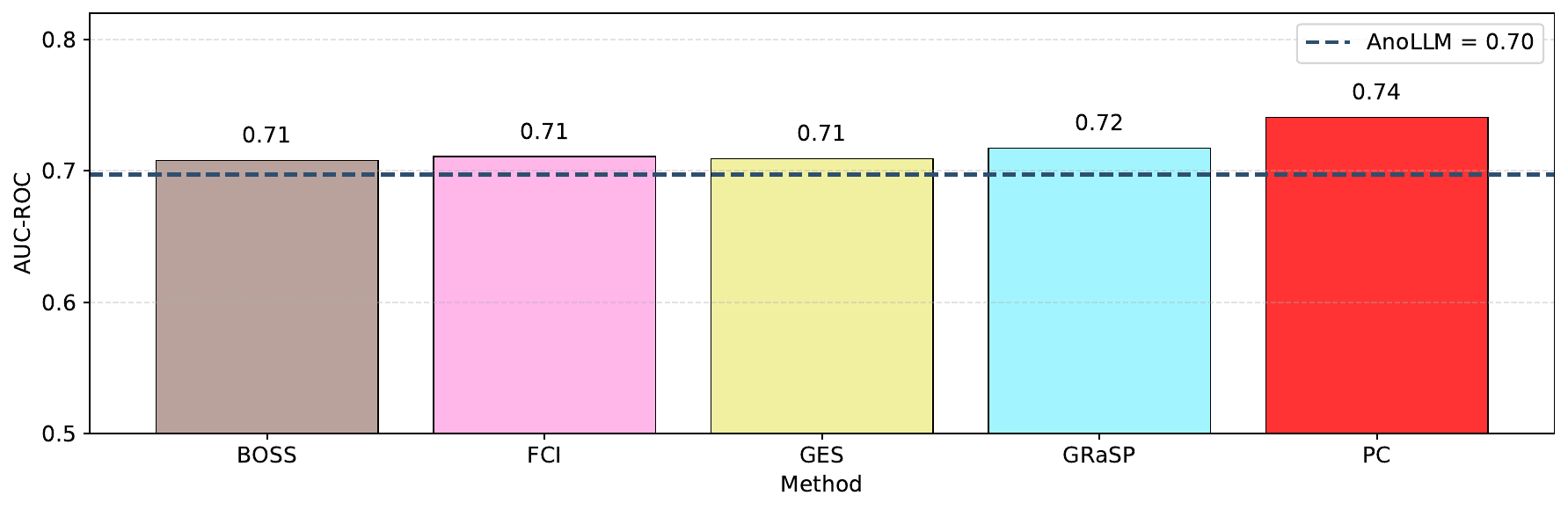}

\caption{\label{fig:Performance-CD}Sensitivity to causal discovery methods
over 12 mixed-type datasets.}
\end{figure}

To further examine the role of causal discovery method in our framework,
we compare PC with four alternative methods: BOSS \cite{andrews2023fast},
FCI \cite{spirtes2013causal}, GES \cite{chickering2002optimal},
and GRaSP \cite{lam2022greedy}. In real-world anomaly detection scenarios,
the ground-truth causal graph is typically unavailable; therefore,
causal structures must be estimated directly from data. Although these
estimated graphs may not perfectly recover the true causal relationships,
the results in Figure \ref{fig:Performance-CD} show that they still
provide useful structural information for constructing causal-aware
scoring logic. While the performance varies slightly across different
causal discovery methods, all variants consistently outperform the
strongest baseline, AnoLLM. This suggests that the performance gain
does not rely on a single causal discovery algorithm, but rather on
the broader benefit of incorporating data-driven causal structure
into the detection process.

\subsubsection{Computational Efficiency}

\begin{figure}[th]
\centering{}\includegraphics[width=0.8\textwidth]{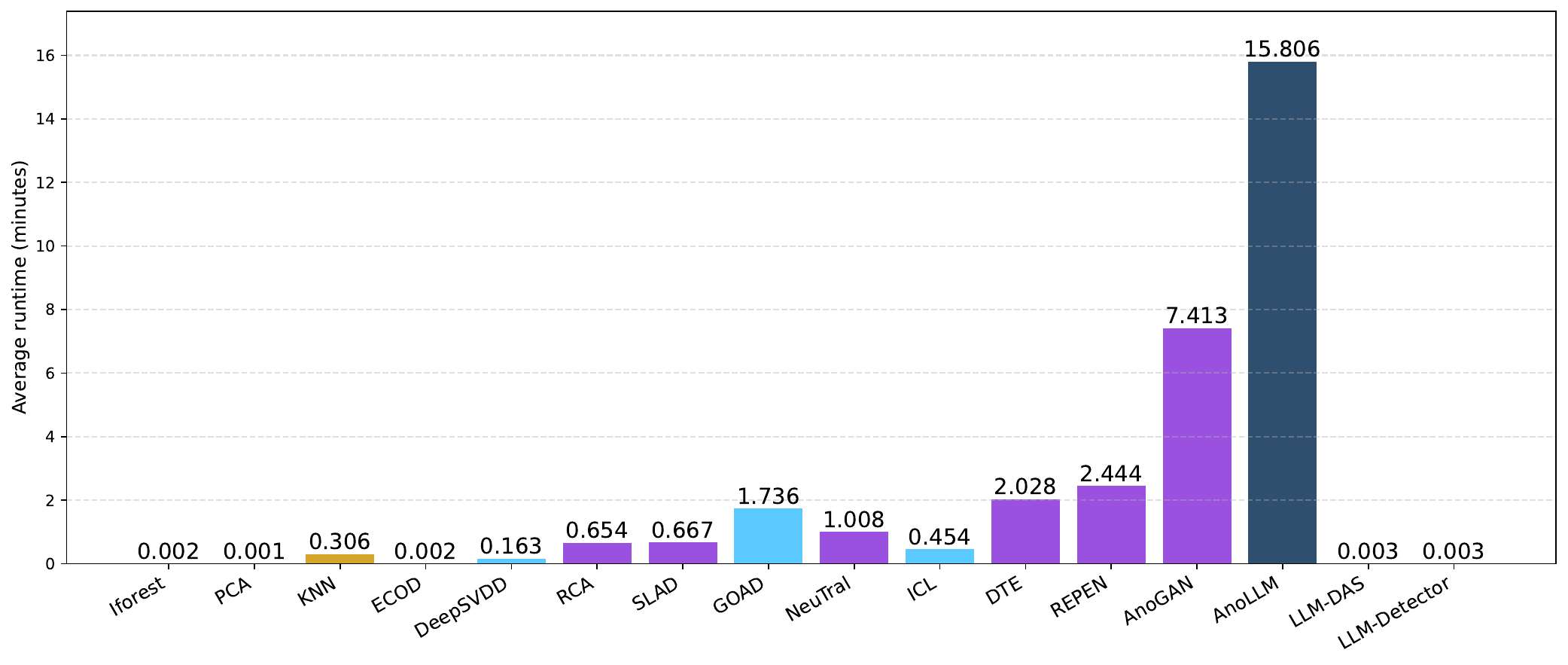}\caption{\label{fig:Computational-efficiency}Computational efficiency comparison
across all 24 datasets.}
\end{figure}

Figure \ref{fig:Computational-efficiency} presents the average runtimes
(\textit{in minutes}) across all 24 datasets on a single GPU: NVIDIA
RTX 4070 Super. LLM-Detector is fast in practice because its test-time
inference reduces to executing a lightweight, deterministic scoring
program. Other methods based on training/fine-tuning e.g., AnoGAN
and AnoLLM require much higher computation.

\section{Conclusion\label{sec:Conclusion}}

We study \textit{unsupervised} anomaly detection in tabular data,
where anomalies often arise from violations of feature dependencies
rather than simple marginal deviations. Unlike other detectors that
rely on reconstruction signals or costly training, our method LLM-Detector
instead treats in-context learning as a structured code synthesis
process, synthesizing an executable anomaly detection engine from
structured normal-state knowledge. The framework first transforms
normal training data into statistical profiles, causal relations,
and distilled prototypes to construct a knowledge-dense prompt. It
then instructs a pre-trained LLM to generate a deterministic scoring
engine that evaluates statistical deviation, structural inconsistency,
and density-based abnormality. Extensive experiments on 24 tabular
datasets demonstrate strong and stable performance, with particularly
clear advantages on mixed-type datasets.

\bibliographystyle{splncs04}
\bibliography{reference}

\appendix
\pagebreak
\section*{Appendix 1: Characteristics of 12 continuous-only datasets}

Table \ref{tab:continuous-data} summarizes the characteristics of
the 12 continuous-only datasets used in our benchmark.

\begin{table}[th]
\centering
\caption{\label{tab:continuous-data}Statistics for 12 continuous-only datasets.}

\begin{tabular}{|l|r|r|r|r|r|}
\hline 
\rowcolor{header_color}Dataset & $N$ & $d$ & $d_{cat}$ & $d_{con}$ & $N_{a}$\tabularnewline
\hline 
\hline 
\textit{Annthyroid} & 7,200 & 6 & 0 & 6 & 534 (7\%)\tabularnewline
\hline 
\rowcolor{even_color}\textit{BreastW} & 699 & 9 & 0 & 9 & 241 (34\%)\tabularnewline
\hline 
\textit{Ecoli} & 336 & 7 & 0 & 7 & 9 (3\%)\tabularnewline
\hline 
\rowcolor{even_color}\textit{Shuttle} & 49,097 & 9 & 0 & 9 & 3,511 (7\%)\tabularnewline
\hline 
\textit{Thyroid} & 3,772 & 6 & 0 & 6 & 93 (2\%)\tabularnewline
\hline 
\rowcolor{even_color}\textit{Wine} & 130 & 13 & 0 & 13 & 71 (55\%)\tabularnewline
\hline 
\textit{WBC} & 223 & 9 & 0 & 9 & 10 (4\%)\tabularnewline
\hline 
\rowcolor{even_color}\textit{Pageblocks} & 5,393 & 10 & 0 & 10 & 510 (9\%)\tabularnewline
\hline 
\textit{Stamps} & 340 & 9 & 0 & 9 & 31 (9\%)\tabularnewline
\hline 
\rowcolor{even_color}\textit{MagicGamma} & 19,020 & 10 & 0 & 10 & 6,688 (35\%)\tabularnewline
\hline 
\textit{Hepatitis} & 80 & 19 & 0 & 19 & 13 (16\%)\tabularnewline
\hline 
\rowcolor{even_color}\textit{Yeast} & 1,484 & 8 & 0 & 8 & 95 (6\%)\tabularnewline
\hline 
\end{tabular}
\end{table}

\section*{Appendix 2: Full results for AUC-ROC}

We reported AUC-ROC as the primary performance metric in the main
paper. Table \ref{tab:auc-all-ta} presents the full results of AUC-ROC.

\begin{table}[th]
\caption{\label{tab:auc-all-ta}AUC-ROC (standard deviation) on each dataset
(\textit{higher is better}).}

\resizebox{\textwidth}{!}{%
\begin{tabular}{|l|c|c|c|c|c|c|c|c|c|c|c|c|c|c|c|c|}
\hline 
\rowcolor{header_color}Dataset & Iforest & PCA & KNN & ECOD & DeepSVDD & RCA & SLAD & GOAD & NeuTral & ICL & DTE & REPEN & AnoGAN & AnoLLM & LLM-DAS & LLM-Detector\tabularnewline
\hline 
\hline 
\rowcolor{even_color}Annthyroid & \begin{cellvarwidth}[t]
\centering
0.9074

(0.011)
\end{cellvarwidth} & \begin{cellvarwidth}[t]
\centering
0.8298

(0.024)
\end{cellvarwidth} & \begin{cellvarwidth}[t]
\centering
0.8088

(0.003)
\end{cellvarwidth} & \begin{cellvarwidth}[t]
\centering
0.7883

(0.002)
\end{cellvarwidth} & \begin{cellvarwidth}[t]
\centering
0.7335

(0.014)
\end{cellvarwidth} & \begin{cellvarwidth}[t]
\centering
0.7160

(0.001)
\end{cellvarwidth} & \begin{cellvarwidth}[t]
\centering
0.7538

(0.008)
\end{cellvarwidth} & \begin{cellvarwidth}[t]
\centering
0.5521

(0.002)
\end{cellvarwidth} & \begin{cellvarwidth}[t]
\centering
0.7766

(0.022)
\end{cellvarwidth} & \begin{cellvarwidth}[t]
\centering
0.8396

(0.019)
\end{cellvarwidth} & \begin{cellvarwidth}[t]
\centering
\textbf{0.9763}

(0.002)
\end{cellvarwidth} & \begin{cellvarwidth}[t]
\centering
0.7316

(0.016)
\end{cellvarwidth} & \begin{cellvarwidth}[t]
\centering
0.6361

(0.155)
\end{cellvarwidth} & \begin{cellvarwidth}[t]
\centering
\uline{0.9153}

(0.001)
\end{cellvarwidth} & \begin{cellvarwidth}[t]
\centering
0.8330

(0.016)
\end{cellvarwidth} & \begin{cellvarwidth}[t]
\centering
0.9117

(0.013)
\end{cellvarwidth}\tabularnewline
\hline 
\end{tabular}}
\end{table}

\section*{Appendix 3: Full results for F1-score}

We report F1-score as a complementary performance metric. Table \ref{tab:f1-all}
reports F1-score and standard deviation for each dataset. On average,
LLM-Detector achieves the best average performance, outperforming
LLM-DAS and AnoLLM by around 5\% and 2\%, respectively, across all
evaluated datasets.

\begin{table}[th]
\caption{\label{tab:f1-all}F1-score (standard deviation) on each dataset (\textit{higher
is better}).}

\resizebox{\textwidth}{!}{%
\begin{tabular}{|l|c|c|c|c|c|c|c|c|c|c|c|c|c|c|c|c|}
\hline 
\rowcolor{header_color}Dataset & Iforest & PCA & KNN & ECOD & DeepSVDD & RCA & SLAD & GOAD & NeuTral & ICL & DTE & REPEN & AnoGAN & AnoLLM & LLM-DAS & LLM-Detector\tabularnewline
\hline 
\hline 
\rowcolor{even_color}Annthyroid & \begin{cellvarwidth}[t]
\centering
0.5512

(0.016)
\end{cellvarwidth} & \begin{cellvarwidth}[t]
\centering
0.4794

(0.025)
\end{cellvarwidth} & \begin{cellvarwidth}[t]
\centering
0.4394

(0.005)
\end{cellvarwidth} & \begin{cellvarwidth}[t]
\centering
0.3882

(0.001)
\end{cellvarwidth} & \begin{cellvarwidth}[t]
\centering
0.4288

(0.025)
\end{cellvarwidth} & \begin{cellvarwidth}[t]
\centering
0.3733

(0.006)
\end{cellvarwidth} & \begin{cellvarwidth}[t]
\centering
0.4051 

(0.003)
\end{cellvarwidth} & \begin{cellvarwidth}[t]
\centering
0.2303

(0.003)
\end{cellvarwidth} & \begin{cellvarwidth}[t]
\centering
0.4244 

(0.013)
\end{cellvarwidth} & \begin{cellvarwidth}[t]
\centering
0.5568

(0.023)
\end{cellvarwidth} & \begin{cellvarwidth}[t]
\centering
\textbf{0.7840}

(0.011)
\end{cellvarwidth} & \begin{cellvarwidth}[t]
\centering
0.3377

(0.014)
\end{cellvarwidth} & \begin{cellvarwidth}[t]
\centering
0.2903

(0.109)
\end{cellvarwidth} & \begin{cellvarwidth}[t]
\centering
\uline{0.6036}

(0.002)
\end{cellvarwidth} & \begin{cellvarwidth}[t]
\centering
0.4644

(0.017)
\end{cellvarwidth} & \begin{cellvarwidth}[t]
\centering
0.4819

(0.096)
\end{cellvarwidth}\tabularnewline
\hline 
\end{tabular}}
\end{table}

\section*{Appendix 4: Full code-generation prompt template}

\begin{figure}[th]
\centering
\includegraphics[width=1\textwidth]{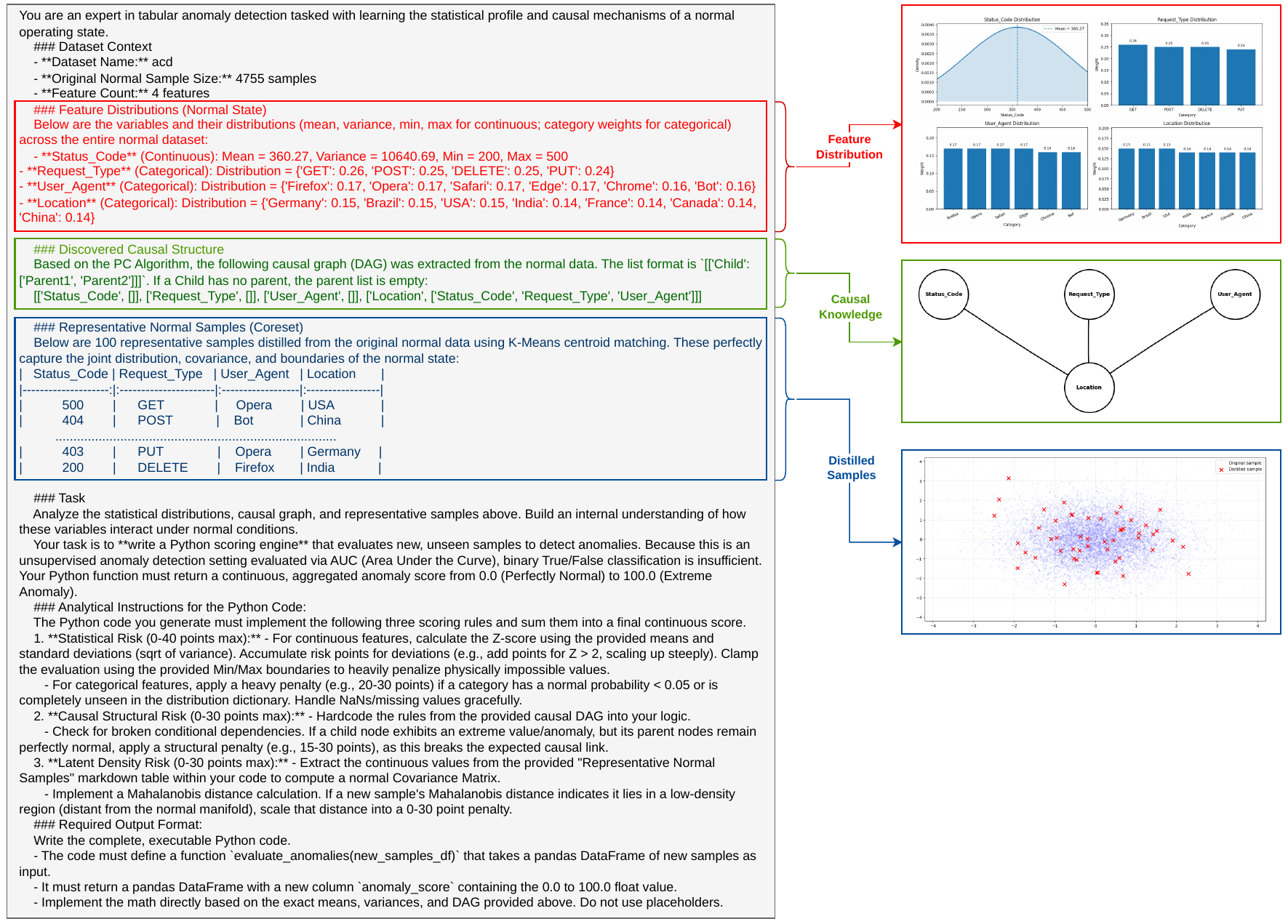}

\caption{\label{fig:Full-code-generation-prompt}Full code-generation prompt
template used in the Knowledge-to-Code stage. The prompt includes
role definition, dataset context, task description, analytical instructions,
and expected output format, while replacing full-dataset injection
with three compact normal-state knowledge components: feature distributions,
causal knowledge, and distilled samples. The distilled sample block
is truncated for space efficiency; the actual setup uses 100 distilled
samples.}
\end{figure}

This appendix provides the complete prompt template used in the Knowledge-to-Code
stage. As shown in Figure \ref{fig:Full-code-generation-prompt},
the prompt is designed to contain the essential components needed
to guide the LLM toward reliable scoring-engine generation. It begins
with a role definition, such as \textquotedblleft You are an expert
in tabular anomaly detection\textquotedblright{} to frame the LLM
as a domain-specific reasoning agent. It then provides dataset context,
including the dataset name, normal sample size, and feature count,
followed by a clear task description, analytical scoring instructions,
and the expected output format. These elements ensure that the LLM
produces a structured anomaly scoring mechanism rather than a generic
explanation.

The main design principle is to replace full-dataset injection with
three compact normal-state knowledge components: feature distributions,
causal knowledge, and distilled samples. The feature distribution
block summarizes marginal normal behavior through numerical statistics
and categorical probabilities. The causal knowledge block encodes
parent-child dependencies discovered from normal data. The distilled
sample block provides representative normal prototypes selected by
K-Means centroid matching. By using these structured summaries instead
of the full normal dataset, the prompt reduces information leakage
risk, remains independent of the original dataset size, and avoids
the token-limit issue of in-context learning. Thus, even for large
datasets, the prompt stays compact while preserving the statistical,
causal, and geometric information required for final anomaly scoring.

\end{document}